\documentclass[11pt]{article}
\usepackage[a4paper,margin=2cm]{geometry}
\usepackage[comma, round]{natbib}
\usepackage{microtype}
\usepackage{subfigure}
\usepackage{booktabs} %

\usepackage{float}
\usepackage{amssymb}
\usepackage{amsthm}
\usepackage{amsmath}
\usepackage{titletoc}
\usepackage{mathrsfs}
\usepackage[colorlinks, allcolors=blue]{hyperref}
\usepackage{amsfonts}
\usepackage{graphicx}
\usepackage{algorithm}
\usepackage{algorithmic}
\usepackage{esint}
\usepackage{bbm}
\usepackage{bm}
\usepackage{mathtools}
\usepackage[shortlabels, inline]{enumitem}
\usepackage{authblk}
\usepackage{todonotes}
\usepackage{color}

\numberwithin{equation}{section}
\numberwithin{figure}{section}

\theoremstyle{plain}
\newtheorem{thm}{Theorem}[section]

\theoremstyle{definition}

\newcommand{\cW}{\mathcal{W}}
\newcommand{\cX}{\mathcal{X}}
\newcommand{\cY}{\mathcal{Y}}
\newcommand{\cZ}{\mathcal{Z}}

\newcommand{\bbR}{\mathbb{R}}

\newcommand{\scrP}{\mathscr{P}}

\newcommand{\1}{\mathbbm{1}}

\newcommand{\W}{\mathcal{W}}

\newcommand{\esssup}{\mathop{\rm{ess}~\sup}}

\newcommand{\E}{E}

\newcommand{\proj}{\operatorname{pj}}

\newcommand{\adv}{\mathrm{adv}}
\newcommand{\sgn}{\mathop{}\mathrm{sgn}}

\def\lb{\mathopen{}\mathclose\bgroup\left}
\def\rb{\aftergroup\egroup\right}

\title{Wasserstein Distributional Adversarial Training for Deep Neural Networks}
\author[1]{Xingjian Bai \thanks{Email: {\tt xbai@mit.edu}, URL: {\tt https://xingjianbai.com}}}
\author[2]{Guangyi He \thanks{Email: {\tt g.he23@imperial.ac.uk}.}}
\author[3]{Yifan Jiang \thanks{Email: {\tt yifan.jiang@maths.ox.ac.uk}, URL: {\tt https://yifanjiang233.github.io}}}
\author[4]{Jan Ob{\l}{\'o}j \thanks{Email: {\tt jan.obloj@maths.ox.ac.uk}, URL: {\tt www.maths.ox.ac.uk/people/jan.obloj}}}
\affil[1]{EECS Department, Massachusetts Institute of Technology}
\affil[2]{Department of Mathematics, Imperial College London}
\affil[3,4]{Mathematical Institute, University of Oxford}
\date{}

\begin{document}

\maketitle

\abstract{
Design of adversarial attacks for deep neural networks, as well as methods of adversarial training against them, are subject of intense research. In this paper, we propose methods to train against distributional attack threats, extending the TRADES method used for pointwise attacks. Our approach leverages recent contributions and relies on sensitivity analysis for Wasserstein distributionally robust optimization problems. We introduce an efficient fine-tuning method which can be deployed on a previously trained model. We test our methods on a range of pre-trained models on RobustBench. These experimental results demonstrate the additional training enhances Wasserstein distributional robustness, while maintaining original levels of pointwise robustness, even for already very successful networks. The improvements are less marked for models pre-trained using huge synthetic datasets of 20-100M images. However, remarkably, sometimes our methods are still able to improve their performance even when trained using only the original training dataset (50k images). 
}

\section{Introduction}

DNNs have been shown to offer powerful solutions in a myriad of context. At the same time, they are often vulnerable and their \emph{robustness}, i.e.,  their ability to perform well under uncertainty, relates to several themes in ML. In this paper, we contribute to this broad field and offer novel understanding of training methods that can deliver robustness to adversarial attacks.

First diagnosed in \citet{BCM+13, GSS15}, it is well known that a crafty attacker can manipulate network's outputs by changing the input images only slightly, and  often in ways which are imperceptible to a human eye. Understanding, and minimizing, such vulnerabilities of trained networks to adversarial attacks (AA), is of key importance for security-sensitive applications. In this active field of work, most research to date has focused on attacks under \emph{pointwise} \(l_{p}\)-bounded image distortions. However, a growing stream of research, including \citet{SJ17,SND18,BHJO23Wasserstein}, frames the problem using Wasserstein distributionally robust optimization (DRO) and considers a \emph{distributional threat model}. We contribute to this literature offering novel training methods for such AA threats. 
Our main contributions can be summarized as follows. \textbf{1)} We extend TRADES training method to account for Wasserstein distributional threat. \textbf{2)} We propose an effective method for fine-tuning pre-trained robust networks. The method aims at maintaining the original robustness against pointwise attacks, while improving robustness against distributional attacks. \textbf{3)} We conduct experiments on a range of successful pre-trained neural networks on RobustBench \citep{CAS+21} and demonstrate effectiveness of the proposed new training method as a fine-tuning device.

\section{Literature Review}

\paragraph{Adversarial Attacks (AA).}
AAs vary between white-box attacks, where the attacker has full knowledge of the neural network and black-box attacks, where they have limited, or no, such knowledge. The latter include, e.g., Zeroth Order Optimization (ZOO) \citep{CZS+17}, Boundary Attack \citep{BRB18}, and Query-limited Attack \citep{IEAL18}. Of more relevance to us are the former attacks, originally focused on the \emph{pointwise} \(l_{p}\)-bounded image distortions, including Fast Gradient Sign Method (FGSM) \citep{GSS15}, Projected Gradient Descent (PGD) \citep{MMS+18}, CW attack \citep{CW17}. A useful benchmark for \(l_{p}\)-robustness of neural networks is provided by AutoAttack \citep{CH20}, an ensemble of various attack methods. More recently, a broader class of \emph{distributional threats} has been considered, see \citet{SJ17,SND18}, and \citet{BHJO23Wasserstein} proposed the first AA associated to such threats using Wasserstein distances. 

\paragraph{Adversarial Training.}
A variety of adversarial training methods, designed to withstand AAs,  have been introduced, including data augmentation, loss regularization, parameter fine-tuning, etc. 
To augment the training data, researchers have used adversarial examples, generated with AAs, see \citep{GSS15,MMS+18,TKP+18}, as well as random modifications, e.g., using generative models such as GANs and diffusion models, see \citep{GRW+21,XSC22,wang2023better}. 
\citet{zhang2019theoretically} consider the trade-off between robustness and accuracy of a neural network via TRADES, a regularized loss. Analogous research includes MART \citep{WZY+20} and SCORE \citep{PLY+22}. 
Loss regularization methods, e.g., adversarial weight perturbation \citep{WXW20}, have been shown to smooth the loss landscape and improve the robustness. 
In addition, various training techniques can be overlaid to improve robustness, including drop-out layers, early stopping and parameter fine-tuning \citet{SWMJ20}.
Among various robust training methods, \citet{SND18,GG22} are directly related to our work as they employ Wasserstein penalization and constraint respectively. In contrast, we use sensitivity analysis of \citet{BDOW21}, to design novel robust training methods. Conceptually close to our work is the adversarial distributional training (ADT) of \citep{DDP+20} with the key difference that we employ Wasserstein distances instead of an entropic regularization.

\paragraph{Distributionally Robust Optimization (DRO).}
DRO provides a unifying language across many papers mentioned above. 
Mathematical formulation involves a min-max problem
\begin{equation}
    \label{eqn-dro}
    \inf_{\theta\in \Theta} \sup_{Q\in\scrP} \E_{Q}[f_{\theta}(Z)],
\end{equation}
where we minimize the worst-case loss over all possible distributions \(Q\in\scrP\). 
In financial economics, such criteria appear in the context of multi-prior preferences, see \citep{GS89,FW15}. We refer to \citep{RM19} for a survey of the DRO. 

Here, we focus on the ambiguity set \(\scrP=B_{\delta}(P)\) given by a ball centered at the reference distribution \(P\) with radius \(\delta\) in the Wasserstein distance, and refer to \citet{GK22} for a discussion of many advantages of this optimal-transport induced distance compared to other, e.g., divergence based, metrics. Importantly for capturing data perturbations, measures with different supports can be close in Wasserstein distance, see \citet{SND18}. Traditional \emph{pointwise} adversarial training can be seen as a special case of Wasserstein DRO (W-DRO) problem, \citet{SJ17}.
More recently, \citet{BLT+22Unified} unified various classical adversarial training methods, such as PGD-AT, TRADES, and MART, under the W-DRO framework. We also mention \citet{volpi2018generalizing} who used W-DRO to improve network performance on unseen data distributions.  

W-DRO, while very compelling, comes at a cost of an optimization over the space of probability measures, often intractable and/or computationally prohibitively expensive. To address this, one can use convex duality to rewrite \eqref{eqn-dro}, changing the $\sup$ to a univariate $\inf$ featuring a transform of $f_\theta$, see \citet{MK18} for the data-driven case, \citet{BM19,BDT20,GK22} for general probability measures and \citet{HHLD22} for a further application with coresets. Another approach, pioneered in \citet{BDOW21}, considers the first order approximation to the original W-DRO problem in the size of the uncertainty radius $\delta$. It was recently employed in \citet{BHJO23Wasserstein} to devise distributional AAs, and we use it here to construct novel robust training methods. 

\paragraph{Fine-tuning.} \label{subsec:finetuning-litreview}
Fine-tuning, in which an already trained model undergoes a limited additional training, often plays an important role in ML applications. 
In particular, a widely adopted strategy in transfer learning is to fine-tune only the final layers of a large pre-trained model, leaving earlier layers frozen. This approach is used not only in robustness tasks, but also in a broad range of scenarios such as domain adaptation \citep{DBLP:conf/cvpr/TzengHSD17} and continual learning \citep{DBLP:journals/nn/ParisiKPKW19}.
Recent work has shown that fine-tuning can substantially improve both standard and adversarial robustness of large pre-trained models \citep{yosinski2014transferable,raghu2019transfusion}. 
By \emph{randomizing} (or re-initializing) the last layer while freezing previous layers, one effectively learns a new ``head'' on top of features that were trained for robustness \citep{kornblith2019better}. 
Such partial re-initialization can avoid catastrophic forgetting of already-acquired robust representations \citep{DBLP:conf/eccv/LiH16}, while still adapting to new threats, including distributional attacks considered in this paper. 
Empirical results suggest that this strategy promotes better generalization in low-data regimes and helps preserve robust features, as noise or domain shifts are mainly absorbed by the newly learned final layer \citep{DBLP:conf/acl/RuderH18}. Furthermore, combining randomization of the last layer with additional regularization (for instance, adversarial weight perturbations or loss smoothing) can reduce overfitting and stabilize fine-tuning \citep{SWMJ20}. 
These observations align with the methods introduced in this paper, where we selectively re-initialize the final layer, or apply minimal random perturbations to all layers, to enhance distributional robustness while preserving core capabilities learned through previous training.

\section{Preliminaries}

\paragraph{Wasserstein Distances.}
Let \((\cZ,d)\) be a Polish space with metric \(d\).
Given two probability measures \(P\) and \(Q\) on \(\cZ\), by \(\Pi(P,Q)\) we denote the set of probability measures on \(\cZ\times\cZ\) whose first margin is \(P\) and second margin is \(Q\).

For \(1\leq p<\infty\), the \(p\)-Wasserstein distance between probability measures \(P\) and \(Q\) on \( \cZ\) is
\begin{equation*}
    \W_{p}(P,Q):=\inf\left\{\E_\pi[d(z_{1},z_{2})^{p}]:\pi\in \Pi(P,Q)\right\}^{1/p}.
\end{equation*}
The $\infty$-Wasserstein distance \(\W_{\infty}\) is given by
\begin{equation*}
    \W_{\infty}(P,Q):=\inf\{\pi\text{--}\esssup d(z_{1},z_{2}):\pi\in \Pi(P,Q)\}.
\end{equation*}

We denote the \(p\)-Wasserstein ball centered at \(P\) with radius \(\delta\) by \(B_{p}(P,\delta)\).
By \(\Pi_{p}(P,\delta)\) we denote the set of couplings 
\(\{\pi: \pi(\cdot\times \cZ)=P(\cdot), \,\E_{\pi}[d(z_{1},z_{2})^p]\leq \delta^{p}\}\).

The Wasserstein distance is a particular case of the optimal transport problem. 
Unlike the KL divergence and total variation distance which are invariant under the change of the metric \(d\),
the Wasserstein distance offers a natural lift of the geometric properties of the underlying space \(\cZ\) to the space of measures on $\cZ$. Wasserstein distances have rich connections to many areas in mathematics, such as PDEs, differential geometry or mean-field problems, and have numerous applications in statistics, finance, and machine learning.

\paragraph{ReDLR Loss.}
Difference of Logits Ratio (DLR) loss introduced in \citet{CH20}, combined with the cross-entropy (CE) loss has been widely recognized as a reliable  method to measure \emph{pointwise} robustness for neural networks. In \citet{BHJO23Wasserstein}, however, it is pointed out that DLR loss is not appropriate for \emph{distributional} threat models.
The authors instead propose ReDLR loss, a rectified version of DLR loss.
It is given by $\mathrm{ReDLR}(z,y)=-(\mathrm{DLR})^{-}(z,y)$, i.e.,
if we write \(z=(z_{1},\dots,z_{m})=f_{\theta}(x)\) for the output of a neural network, and \(z_{(1)}\geq\dots\geq z_{(m)}\) are the order statistics of \(z\), then

\begin{equation*}
\label{eqn-redlr}
    \mathrm{ReDLR}(z,y)=\left\{\begin{aligned}
        -\frac{z_{y}-z_{(2)}}{z_{(1)}-z_{(3)}}, & \quad\text{if } z_{y}=z_{(1)}, \\
        0,                                                  & \quad\text{else.}
    \end{aligned}\right.
\end{equation*}
In a key difference with the DLR, the ReDLR loss leaves misclassified images unaffected. This allows distributional attacker to ``save" their budget for images which actually need to be attacked. 
Intuitively, such attacker will perturb more aggressively those images far from the decision boundary. The resulting ReDLR \emph{distributional} adversarial attack was shown to be effective in \citet{BHJO23Wasserstein} across a wide range of models on RobustBench \citep{CAS+21}. These results underscored the need to develop training methods to defend against distributional AAs.

\section{Wasserstein Distributional Adversarial Training}
\paragraph{W-DRO Reformulation.} 

Let \(\cX=[0,1]^{n}\) be the feature space of an image, and \(\cY=\{1,\dots,m\}\) the label space.
We equip \(\cZ=\cX\times \cY\) with a metric 
\begin{equation}
    \label{eqn-d}
    d((x_{1},y_{1}),(x_{2},y_{2}))=\|x_{1}-x_{2}\|_{\infty}+\infty\1_{\{y_{1}\neq y_{2}\}}.
\end{equation}
An image is interpreted as a pair \((x,y)\in \cX\times \cY\).
A neural network is a parameterized map \(f_{\theta}:\cX\to\bbR^{m}\).

Adversarial training, including influential formulations in  \citet{MMS+18,zhang2019theoretically}, is fundamentally a min-max problem: an inner maximization picks an adversarial perturbation of the data and the outer minimization of the loss learns the best network parameter $\theta$. 
Recently, \citet{BLT+22Unified} proposed a more general criterion which includes as special cases many previous training methods such as MART and TRADES. Their \emph{unified distributional robustness} can be re-cast equivalently as an \emph{extended} W-DRO formulation using couplings:
\begin{equation}
 \label{eqn-advl-gen}
    \inf_{\theta\in\Theta}\sup_{\pi\in \Pi_{p}(P, \delta)}\E_{\pi}[J_{\theta}(x,x',y)].
\end{equation}
We are mainly interested in \(p\in\{2,\infty\}\) and note that \(p=\infty\) corresponds to the \emph{pointwise} threat model. The case \(p=2\) is the \emph{distributional} threat model we consider in this paper.

\paragraph{W-DRO Sensitivity.}
To back propagate \(\theta\) it is essential to understand the inner maximization problem denoted by
\begin{equation*}
    V(\delta)=\sup_{\pi\in \Pi_2(P,\delta)}\E_{\pi}[J_{\theta}(x,x',y)].%
\end{equation*}
This is an infinite-dimensional optimization problem over probability measures. Using convex duality, in the spirit of \citet{BM19}, it can be recast as another optimization problem but at a cost of introducing a new Lagrangian multiplier \(\lambda\). Training with a fixed $\lambda$ is then feasible, see \citet{SND18}, and such a fixed \(\lambda\) may be optimal for \emph{some} budget \(\delta\), but there is no guarantee that a given choice of \(\lambda\) is even near optimal for a \emph{given} budget \(\delta\).

In \citet{BHJO23Wasserstein}, a sensitivity approach is adapted to address the above issue.
Instead of finding the exact value of \(V(\delta)\), a first order approximation to \(V(\delta)\) in the size of attack budget $\delta$ is derived.
The following result is adapted from \citet[Theorem 2.2]{BDOW21} and \citet[Theorem 4.1]{BHJO23Wasserstein} and their proofs.
\begin{thm}
    \label{thm-u}
    Assume the map \((x,x',y)\mapsto J_{\theta}(x,x',y)\) is Lipschitz.
    Then the following first order approximations hold:
    \begin{enumerate}[(i)]
        \item \label{prop-u:i}
              \(
              V(\delta)=V(0)+\delta\Upsilon+o(\delta),
              \)
              where
              \begin{equation*}
                  \Upsilon=\Bigl(\E_{P}\|\nabla_{x'}J_{\theta}(x,x,y)\|_{1}^{2}\Bigr)^{1/2}.
              \end{equation*}
        \item \label{prop-u:ii}
              \(V(\delta)=\E_{\pi_{\delta}}[J_{\theta}(x,x,y)]+o(\delta),\)
              where
              \begin{equation*}
                  \pi_{\delta}=\Bigl[(x,y)\mapsto (x,y, x+\delta T(x), y)\Bigr]_{\#}P,
              \end{equation*}
              and 
              \begin{equation*}
                T(x)=\Upsilon^{-1} \sgn(\nabla_{x'}J_{\theta}(x,x,y))\|\nabla_{x'}J_{\theta}(x,x,y)\|_{1}.
            \end{equation*}
    \end{enumerate}
\end{thm}
Although We focus on the \(l_{\infty}\)-distance between images and the \(\cW_2\)-distance between probability measures, the above results extend to a more general \((l_r,\cW_p)\) setting with \(l,p>1\). 
We remark that our assumption is weaker than the $\mathsf{L}$-smoothness assumption encountered in the literature which requires Lipschitz continuity of gradients of $J_\theta$, see for example \citet[Assumption B]{SND18} and \citet[Assumptions 1 \& 2]{volpi2018generalizing}.

This result led to a natural first order distributional attack given in \citet{BHJO23Wasserstein}.

\section{Training Methods}\label{sec:trainmethod}
As highlighted above, taking \(J_{\theta}(x,x',y)=L(f_{\theta}(x),y)+\beta L(f_{\theta}(x),f_{\theta}(x'))\) in \eqref{eqn-advl-gen} and $p=\infty$ recovers the TRADES loss function on \citet{zhang2019theoretically}. 
The training is split into two steps: adversarial attack on the input data \(x\), and optimization over the network parameter \(\theta\).
To consider a Wasserstein distributional threat, we propose to replace the inner \(\Pi_\infty(P,\delta)\) with \(\Pi_2(P,\delta)\).
We consider 
\begin{equation}\label{eq:WTrades}
    \inf_{\theta\in\Theta}\Bigl\{\E_{P}\bigl[ L(f_{\theta}(x),y)\bigr]+\beta\sup_{\pi \in  \Pi_{2}(P,\delta)}E_{\pi}\bigl[\tilde L(f_{\theta}(x),f_{\theta}(x'))\bigr]\Bigr\},
\end{equation}
where \(\beta\) is a hyperparameter leveraging the trade-off between network  clean accuracy and distributional robustness. 

In \citet{zhang2019theoretically}, the cross-entropy loss is used for both the AA step and the optimization step.
However, in face of a distributional threat, we need to separate the loss for the adversarial attack and the one for neural network learning.
Specifically, we take $\tilde L$ as the ReDLR loss for the AA step when computing the adversarial $\pi$.
It has been verified that ReDLR gives a more effective \emph{distributional}  adversarial  attack compared to the one using KL divergence or DLR loss, see \citet{BHJO23Wasserstein}.

On the other hand, ReDLR is not appropriate  for network parameter optimization. This is due to the fact that ReDLR loss is flat when the image is misspecified. We keep \(L\) and $\tilde L$ as the cross-entropy for this step. We refer to Algorithm \ref{alg:train} for details. 

\paragraph{Details of the AA Step.}
For the AA step we apply the W-PGD-20 attack developed in \citet{BHJO23Wasserstein} to 
\begin{equation}
    \sup_{\pi \in  \Pi_{2}(P,\delta)}\E_{\pi}\bigl[\mathrm{ReDLR}(f_{\theta}(x),f_{\theta}(x'))\bigr].
\end{equation}
Specifically, the W-PGD step is given by
\begin{equation}\label{eq:WPDGstep}
    x \leftarrow \proj_{\delta}(x+\alpha \Upsilon^{-1}  \sgn(\nabla_{x}J_{\theta}(x,y))\|\nabla_{x}J_{\theta}(x,y)\|_{1}),
\end{equation}
where \(\alpha=\delta/20\) is the step size and \(\proj_{\delta}\) is a projection which ensures the attacked data stayed within the Wasserstein Ball \(B_{p}(P,\delta)\). 
We point out that \(\Upsilon\) is a global quantity depending on the whole training set and has to be recalculated after every update of the network \(\theta\). Moreover, the projection is also a global operator which depends on the data outside the current batch.
This, in principle, leads to a huge computation burden for the training.
We address these two issues by introducing a W-PGD-budget method.
At the start of each batch, we randomly sample $10\%$ of the whole training set to estimate the current \(\Upsilon\) and apply a moving-average update given by 
\begin{equation}
    \Upsilon \leftarrow \Upsilon + \eta (\hat{\Upsilon}-\Upsilon).
\end{equation}
For each batch, we first calculate the budget \(\delta_{B}\) of the current batch given by 
\begin{equation}
    \delta_{B}=\delta\frac{\Upsilon_{B}}{\Upsilon} =\frac{\delta}{\Upsilon}  \Bigl(\E_{P_{B}}\|\nabla_{x}J_{\theta}(x,y)\|_{1}^{2}\Bigr)^{1/2},
\end{equation}
where \(P_{B}\) is the empirical distribution of the current batch.
We then apply a W-PGD-20 attack locally on the current batch with budget \(\delta_{B}\) which is more computationally feasible.
It is clear that if \(\theta\) was frozen during the training, the budget \(\delta_{B}\) on each batch would lead to an admissible attack on the whole training set with budget \(\delta\).
We refer to Algorithm \ref{alg:train} for the detailed training procedure.

\begin{algorithm}[tb]
   \caption{Wasserstein Distributional Adversarial Training with W-PGD-Budget Attack}
   \label{alg:train}
\begin{algorithmic}
   \STATE {\bfseries Input:} training dataset $D$, hyperparameters $\eta$, $\delta$, $\tau$, $\beta$
   \REPEAT
       \STATE Sample $D$ and  estimate the sensitivity $\hat{\Upsilon}$.
        \STATE Update \(\Upsilon\)
        \[    \Upsilon \leftarrow \Upsilon + \eta (\hat{\Upsilon}-\Upsilon).\]       
   \STATE Generate a minibatch $B$ from $D$
    \STATE Calculate the budget $\delta_{B}$ for the minibatch
    \[    \delta_{B} \leftarrow\frac{\delta}{\Upsilon} \Bigl( \frac{1}{|B|}\sum_{(x,y)\in B}\|\nabla_{x'}J_{\theta}(x,x,y)\|_{1}^{2}\Bigr)^{1/2},\]
    where \(J_{\theta}(x,x',y)=\mathrm{ReDLR}(f_{\theta}(x),f_{\theta}(x'))\).
    \STATE Apply W-PGD-Budget attack  with budget $\delta_{B}$
   \[B_{\adv}\leftarrow \text{W-PGD}(B,\delta_B).\]
   \STATE Update the network parameter $\theta$ by SGD.
    \[\theta\leftarrow \theta - \tau \frac{1}{|B|}\sum_{(x,x_{\adv},y)\in (B,B_{\adv})}\nabla_{\theta}J_{\theta}(x,x_{\adv},y),\]
    where  \[J_{\theta}(x,x',y)= \mathrm{CE}(f_{\theta}(x),y) +\beta \mathrm{CE}(f_{\theta}(x),F_{\theta}(x')).\]

   \UNTIL{the end condition}
\end{algorithmic}
\end{algorithm}

\paragraph{Training vs Fine-tuning.} 
In principle we could use \eqref{eq:WTrades} to train a DNN from scratch. However, in practice, users are likely to care both about the pointwise and distributional robustness. It would be thus natural to consider an objective combining the two 
\begin{align}
    \inf_{\theta\in\Theta}\Bigl\{&\E_{P}\bigl[ L(f_{\theta}(x),y)\bigr]
    +\beta\sup_{\pi \in  \Pi_{2}(P,\delta)}E_{\pi}\bigl[L(f_{\theta}(x),f_{\theta}(x'))\bigr]\nonumber \\
    &+ \gamma\sup_{\pi \in  \Pi_{\infty}(P,\delta)}E_{\pi}\bigl[L(f_{\theta}(x),f_{\theta}(x'))\bigr]
    \Bigr\},\label{eq:combinedloss}
\end{align}
where $\beta$ and $\gamma$ are hyperparameters which need to be carefully balanced to adjust the two sources of adversarial robustness. Such training would involve a very significant computational effort and we leave it for future research. 

Instead, as argued above, our motivation was to consider \eqref{eq:WTrades} as a fine-tuning method. We are interested in taking pre-trained DNNs which exhibit good, or very good, robustness against pointwise AAs and perform limited additional fine-tuning training using \eqref{eq:WTrades} to improve their robustness against distributional attacks whilst maintaining their pointwise robustness. As is common in the fine-tuning literature, we will perturb our pre-trained model, either by randomizing the last layer or by adding small noise to the whole $\theta$, before performing additional limited training of up to $20$ epochs. 

\section{Experiments}

In this section, we present the experiment setup based on our proposed training methods.
We consider a distributional threat model \((\cW_{2},l_{\infty})\) with size \(\delta=8/255\).
The adversarial accuracy reported in this paper is evaluated under 20-step W-PGD attack on the test set with CE loss for the pointwise threat and with ReDLR loss for the distributional threat. 

\paragraph{Dataset.}
We use the CIFAR-10 dataset, which consists of 60,000 32x32 color images in 10 classes, with 6,000 images per class.
There are 50,000 training images and 10,000 test images.
We further split the  50,000 training images into a training set and a validation set. 
The training set contains 40,000 images and the validation set contains 10,000 images.
Throughout the training process, we only draw minibatch from the training set to update the network and use the validation set to track the distributional robustness of the network.
We report the networks' performance according to their best performing epoch on the validation set.

\paragraph{Configuration of the Training.}
The training involves the use of five pre-trained networks from RobustBench which are introduced in details below. 
We adopt a distributional TRADES framework introduced previously with \(\beta=6\).
CE is employed as the loss function for network training, while ReDLR loss is used for distributional adversarial attacks.
We employ a 20 step W-PGD adversarial attack to generate new adversarial images for each minibatch.
The sensitivity \(\Upsilon\) is estimated based on a 10\% random sampling from the training set and is updated in every minibatch.
The update rate \(\eta\) for \(\Upsilon\) is set to 0.1 or 1. 
We take a vanilla SGD optimizer whose learning rate is chosen as either 0.1 or 0.01, and  train the network for 20 epochs.
The batch size is set as 512.
To mitigate overfitting, we introduce random perturbations to the pre-trained networks. Specifically, we either randomly initialize the last layer and restrict updates to this layer or apply a small perturbation to the entire network and train all layers.

\paragraph{Pre-trained Neural Networks.}
From RobustBench, we choose 5 typical successful neural networks  which are robust against the pointwise threat model: \citet{zhang2019theoretically}, \citet{chen2024data}, \citet{gowal2020uncovering} , \citet{cui2023decoupled}, \citet{wang2023better}.
Due to computational constraints, we mainly focus on the networks with relatively small architectures: WideResNet 28-10 and WideResNet 34-10.
In theory, our method is applicable to networks with other architectures.
We point out that the first two were trained merely on the original training set, while the rest used either an extra dataset from ImageNet or a synthetic dataset generated from CIFAR-10.

\citet{zhang2019theoretically} is the original TRADES network trained on CIFAR-10 with 100 epochs. 
\citet{chen2024data} is a recent leading network on RobustBench with architecture WideResNet 28-10. 
It was trained on the original CIFAR-10 with 115 epochs.
\citet{gowal2020uncovering} is a network with WideResNet 28-10 architecture and was trained on an external dataset extracted from  the tiny images dataset 80M-Ti with 400 epochs.
\citet{cui2023decoupled} and \citet{wang2023better} are two top networks with WideResNet 28-10 architecture on RobustBench.
\citet{cui2023decoupled} was trained on a 20M synthetic dataset generated from CIFAR-10 and with 200 epochs.
\citet{wang2023better} was also trained on a 20M synthetic dataset with 2400 epochs.

The scale of the pre-training efforts puts into perspective our fine-tuning which is done, we stress, using only the original CIFAR-10 dataset at at most $20$ epochs. 

In Table \ref{tab-networks}, we present an overview and baseline of these pre-trained networks.  
We remark that both pointwise and distributional robustness are evaluated under 20-step PGD attack on the test set with budget \(\delta=8/255\).

\begin{table*}[tb]
\caption{Overview of the pre-trained networks.}
\label{tab-networks}
\begin{center}
    \scriptsize
\begin{tabular}{lccccc}
    \toprule
Networks  &  Architectures & Extra Dataset & Clean Acc & \(\cW_{\infty}\) Adversarial Acc & \(\cW_{2}\) Adversarial Acc
\\  
\midrule 
\citet{zhang2019theoretically}       &WDR 34-10 & no &84.92 & 57.05 &42.99 \\
\citet{chen2024data}           &WDR 34-10 & no & 86.54 & 60.14 &43.76 \\
\citet{gowal2020uncovering}   &WDR 28-10 &external data  & 89.48 &66.44    &50.99  \\
\citet{cui2023decoupled} &WDR 28-10 & 20M synthetic data& 92.16 &70.92 &53.16\\
\citet{wang2023better} &WDR 28-10 & 20M synthetic data & 92.44 &70.62 &52.14\\
\bottomrule
\end{tabular}

\end{center}
\end{table*}

\section{Experiment Results}
\begin{figure}[ht]
    \vskip 0.2in
    \begin{center}
    \centerline{\includegraphics[width=0.5\textwidth]{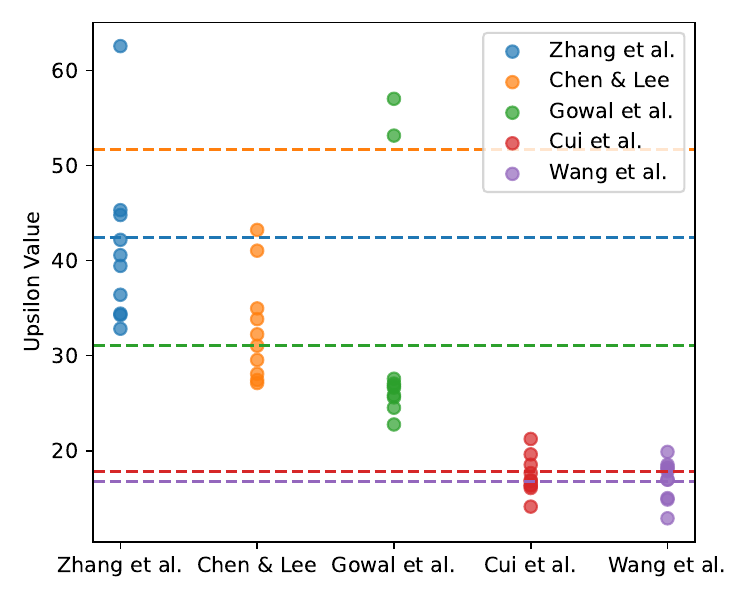}}
    \caption{The estimated \(\Upsilon\) from 10\% of the training set and its reference value for 5 pre-trained networks.}
    \label{upsilon-sample}
    \end{center}
    \vskip -0.2in
    \end{figure}

\paragraph{Effectiveness of the W-PGD-Budget Method.}
In Figure \ref{upsilon-sample}, we present the estimated \(\Upsilon\) from 10\% of the training set and its reference value for 5 pre-trained networks.
For each network, we sampled 10\% of the training set 10 times independently and calculated the \(\Upsilon\) for each sample.
The reference value is calculated based on the whole training set and is marked in dotted lines.
We notice that for more robust networks \citet{wang2023better} and \citet{cui2023decoupled}, the estimated \(\Upsilon\) is closer to the reference value.

\begin{table*}[tb]
    \caption{Performance of W-PGD-Budget-20 on different pre-trained networks.\label{tb:wpgdbudget}}
    \label{tab-budget}
    \begin{center}
    \begin{tabular}{lccc}
        \toprule
    Networks  &  W-PGD-Budget & \(\cW_{\infty}\) Adversarial Acc & \(\cW_{2}\) Adversarial Acc
    \\  
    \midrule 
    \citet{zhang2019theoretically}     &63.99 & 70.61 &57.56\\
    \citet{chen2024data}          & 68.61  &76.27  &61.32\\
    \citet{gowal2020uncovering}   & 70.16& 80.58 & 67.02\\
    \citet{cui2023decoupled} &76.57&90.49 &76.54\\
    \citet{wang2023better} &78.19 & 90.64&77.21     \\
    \bottomrule
    \end{tabular}
    \end{center}
    \end{table*}

In Table \ref{tb:wpgdbudget}, we report the comparison of \(\cW_{\infty}\) adversarial attack, \(\cW_{2}\) adversarial attack,  the budget \(\cW_{2}\) adversarial attack on 5 pre-trained networks.
We observe that, as expected, the budget \(\cW_{2}\) adversarial attack is slightly less effective than the \(\cW_{2}\) adversarial attack, but is much stronger than the \(\cW_{\infty}\) attack.

\paragraph{A Case study.}
We focus on the pre-trained network from \citet{zhang2019theoretically} which has 57.05\% \(\cW_{\infty}\) adversarial accuracy and 42.99\% \(\cW_{2}\) adversarial accuracy.
In Table \ref{tab-last}, we report the result of fine-tuning the last layer of the network with different learning rate \(\tau=0.1, 0.01\) and update rate \(\eta=0.1,1\).
In Table \ref{tab-whole}, we report the result of perturbing the whole network with a small noise and training all the layers.
All the results are chosen for the epoch with the best \(\cW_{2}\) performance on the validation set.
We observe that only fine-tuning the last layer does not improve the network's robustness.
However, perturbing the network slightly and training all the layers can improve  both \emph{pointwise} and \emph{distributional}  robustness for \citet{zhang2019theoretically}.

\begin{figure}[ht]
        \vskip 0.2in
        \begin{center}
        \centerline{\includegraphics[width=0.5\textwidth]{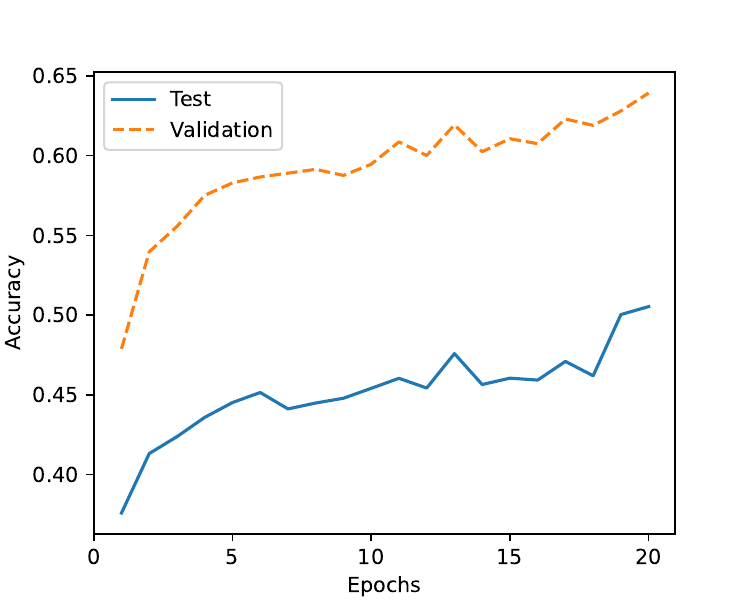}}
        \caption{Performance of Zhang et al. (2019) under \(\cW_{2}\) adversarial attack on the test set (solid) and the validation set (dashed) along fine-tuning.}
        \label{w2}
        \end{center}
        \vskip -0.2in
        \end{figure}

\begin{figure}[ht]
        \vskip 0.2in
        \begin{center}
        \centerline{\includegraphics[width=0.5\textwidth]{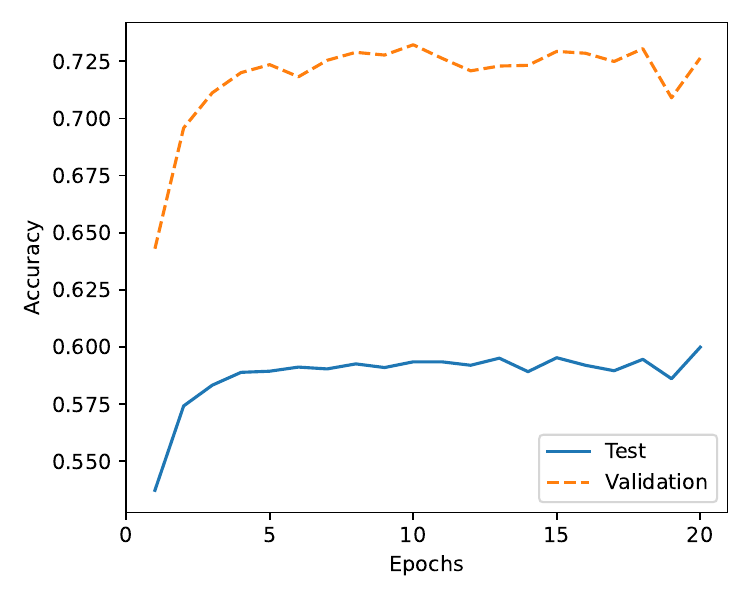}}
        \caption{Performance of Zhang et al. (2019) under \(\cW_{\infty}\) adversarial attack on the test set (solid) and the validation set (dashed) along fine-tuning.}
        \label{winf}
        \end{center}
        \vskip -0.2in
        \end{figure}

 The best performance is achieved with \(\tau=0.01\) and \(\eta=1\) for the whole network fine-tuning.
 In Figure \ref{w2} and Figure \ref{winf}, we show the   \(\cW_{2}\) and \(\cW_{\infty}\) adversarial accuracies of \citet{zhang2019theoretically} along training under this configuration.
 The solid line is the accuracy on the test set, and the dashed line is the accuracy on the validation set.
This gives 59.99\% \(\cW_{\infty}\) adversarial accuracy and 50.53\% \(\cW_{2}\) adversarial accuracy which improves the original pre-trained network by 2.95\% and 7.54\% respectively.

    \begin{figure}[ht]
        \vskip 0.2in
        \begin{center}
        \centerline{\includegraphics[width=0.5\textwidth]{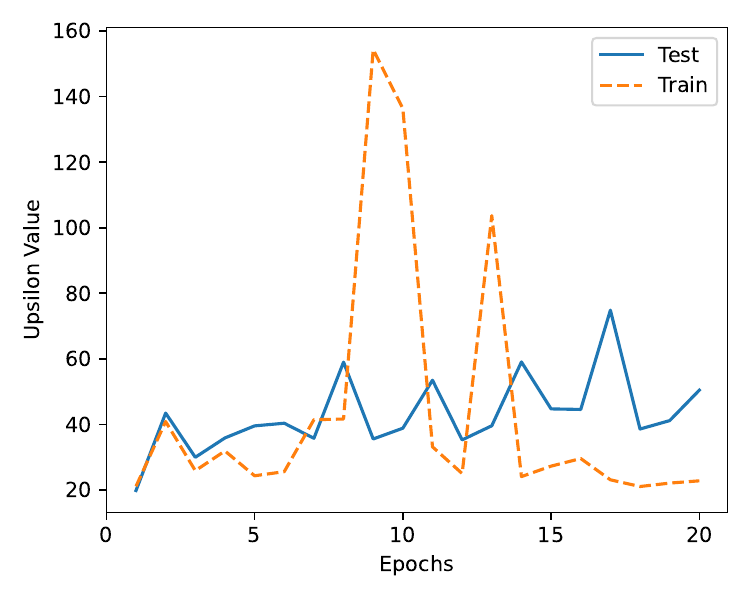}}
        \caption{\(\Upsilon\) of Zhang et al. (2019) along fine-tuning evaluated on the test set (solid) and the training set (dashed).}
        \label{upsilon}
        \end{center}
        \vskip -0.2in
        \end{figure}

        \begin{table*}[tb]
            \caption{Performance of Zhang et al. (2019) under different configurations. We randomized the last layer and fine-tuned the network by only training this layer.}
            \label{tab-last}
            \begin{center}
            \begin{tabular}{lcccc}
                \toprule
              &  \(\tau=0.1, \eta=0.1\)& \(\tau=0.01, \eta=0.1\)& \(\tau=0.1, \eta=1\) & \(\tau=0.01, \eta=1\)
            \\  
            \midrule 
            Clean Acc&  83.89 & 83.86 & 83.56 & 83.63 \\ 
            \(\cW_{\infty}\) Adversarial Acc & 53.47 & 52.78 & 54.25 & 54.20\\
            \(\cW_{2}\) Adversarial Acc& 39.53 & 37.91 & 41.76 & 39.08\\
            \bottomrule
            \end{tabular}
            \end{center}
    \end{table*}

            \begin{table*}[tb]
            \caption{Performance of Zhang et al. (2019) under different configurations. We perturbed the whole network with a small noise and trained all the layers.}
            \label{tab-whole}
            \begin{center}
            \begin{tabular}{lcccc}
                \toprule
              &  \(\tau=0.1, \eta=0.1\)& \(\tau=0.01, \eta=0.1\)& \(\tau=0.1, \eta=1\) & \(\tau=0.01, \eta=1\)
            \\  
            \midrule 
            Clean Acc& 83.38 & 85.15 & 82.89 & 83.71 \\ 
            \(\cW_{\infty}\) Adversarial Acc & 58.50 & 58.63& 57.55 &\textbf{59.99}\\
            \(\cW_{2}\) Adversarial Acc&  47.91 & 45.78 & 45.20 & \textbf{50.53}\\
            \bottomrule
            \end{tabular}
            \end{center}
    \end{table*}

\begin{table*}[tbh!]
    \caption{ Summary of the best fine-tuning performance for the five pre-trained networks. The change from the baseline performance is indicated in parentheses. \label{tb:results}}
    \label{tab-ft}
    \begin{center}
    \begin{tabular}{lccc}
        \toprule
    Networks  &  Clean Acc & \(\cW_{\infty}\) Adversarial Acc & \(\cW_{2}\) Adversarial Acc
    \\  
    \midrule 
    \citet{zhang2019theoretically}     &83.71 & 59.99 (+2.95) &50.53 (+7.54) \\
    \citet{chen2024data}          & 85.44  & 62.12 (+1.98) &53.42 (+9.66)\\
    \citet{gowal2020uncovering}   & 85.93 & 63.39 (-3.05)  & 52.14 (+1.15)\\
    \citet{cui2023decoupled} &88.88 &68.71 (-2.21) &58.02 (+4.86)\\
    \citet{wang2023better} & 91.45 & 69.19 (-1.43)&55.93 (+3.79)\\
    \bottomrule
    \end{tabular}
    \end{center}
    \end{table*}
\section{Limitations}
    \paragraph{Fine-tuning Other Pre-trained Networks.}
We summarize the best fine-tuning performance for the 5 pre-trained networks in Table \ref{tab-ft}, and  provide a complete report in Appendix \ref{ap-a}.
Remarkably, we improve the distributional adversarial accuracy for all five pretrained networks by fine-tuning on the original training set and with 20 epochs of training only.
For those networks only trained on the original CIFAR-10, we improve not only \(\cW_{2}\) adversarial accuracy but also \(\cW_{\infty}\) adversarial accuracy. In addition, this performance was achieved without any extensive hyperparameters optimization. 

The detailed best configuration for  each network is as follows.
For \citet{zhang2019theoretically,gowal2020uncovering,chen2024data}, fine-tuning the whole network with \(\tau=0.01\) and \(\eta=1\) gives the best performance;
for \citet{cui2023decoupled,wang2023better}, fine-tuning the last layer of the network with \(\tau=0.1\) and \(\eta=1\) gives the best performance.
In Figure \ref{upsilon}, we report \(\Upsilon\) of the network along the fine-tuning process (with $\eta=1$) for our case study on \citet{zhang2019theoretically}. We observe that the training curve of \(\Upsilon\) can be fairly volatile.
We introduce the update rate \(\eta\) intended to decrease the oscillation of \(\Upsilon\) and the variance appeared from the sampling.
However, from the empirical results, we find that using the current estimate of \(\Upsilon\) \((\eta=1)\) always outperforms a moving average update \((\eta=0.1)\).
For fine-tuning the last layer, a larger learning rate \(\tau=0.1\) is preferred while for fine-tuning the whole network, a smaller learning rate \(\tau=0.01\) is preferred.

As pointed out above, \(\Upsilon\) is a global quantity depending on the whole training set and it needs to be recalculated after every update of the network \(\theta\). Similarly, the projection is also a global operator which depends on the data outside the current batch. In section \ref{sec:trainmethod} we proposed a workaround involving using a $10\%$ sample of the dataset, which exhibited good performance, as documented above. Whilst this was feasible when working with the actual CIFAR-10 dataset, it would no longer be so when using much larger synthetic datasets. Adapting our methods would likely involve subsampling. 

More broadly, we do not expect our methods to improve all pre-trained models, particularly those trained using massive synthetic datasets. Indeed, Table \ref{tb:results} shows a small decrease in $\cW_\infty$ (pointwise) robustness for \citet{gowal2020uncovering,cui2023decoupled,wang2023better} which were pre-trained using huge additional datasets but fine-tuned using only the original CIFAR-10 dataset. To avoid this effect, it might be necessary to consider training from scratch using a combined objective as in \eqref{eq:combinedloss}. This would be much more involved computationally than the solutions considered in this work.

\section*{Acknowledgements}
GH's research is supported by the Department of Mathematics at Imperial College London through the Roth Scholarship. YJ's research is supported by the EPSRC Centre for Doctoral Training in Mathematics of Random Systems: Analysis, Modelling and Simulation (EP/S023925/1). JO gratefully acknowledges the support of St John's College, Oxford. 

\bibliography{ref.bib}

\newpage
\appendix
\onecolumn
\section{Results on other pre-trained networks.}
\label{ap-a}
We report other pre-trained networks considered in the paper under different training configuration.

   \begin{table*}[htbp]
            \caption{Performance of Gowal et al. (2020) under different configurations. We randomized the last layer and fine-tuned the network by only training this layer.}
            \begin{center}
            \begin{tabular}{lcccc}
                \toprule
              &  \(\tau=0.1, \eta=0.1\)& \(\tau=0.01, \eta=0.1\)& \(\tau=0.1, \eta=1\) & \(\tau=0.01, \eta=1\)
            \\  
            \midrule 
            Clean Acc&  89.68 & 89.75 & 90.02 & 89.43 \\ 
            \(\cW_{\infty}\) Adversarial Acc & 64.20 & 63.95 & 64.81 & 63.56\\
            \(\cW_{2}\) Adversarial Acc& 50.76 & 48.42 & 50.65 & 48.05\\
            \bottomrule
            \end{tabular}
            \end{center}
    \end{table*}

            \begin{table*}[htbp]
            \caption{Performance of  Gowal et al. (2020) under different configurations. We perturbed the whole network with a small noise and trained all the layers.}
            \begin{center}
            \begin{tabular}{lcccc}
                \toprule
              &  \(\tau=0.1, \eta=0.1\)& \(\tau=0.01, \eta=0.1\)& \(\tau=0.1, \eta=1\) & \(\tau=0.01, \eta=1\)
            \\  
            \midrule 
            Clean Acc& 81.04 & 88.09 & 78.15 & 85.93 \\ 
            \(\cW_{\infty}\) Adversarial Acc & 51.97 & 63.05& 50.12 &\textbf{63.39}\\
            \(\cW_{2}\) Adversarial Acc&  39.31 & 49.93 & 40.13 & \textbf{52.14}\\
            \bottomrule
            \end{tabular}
            \end{center}
    \end{table*}

       \begin{table*}[htbp]
            \caption{Performance of Cui et al. (2023) under different configurations. We randomized the last layer and fine-tuned the network by only training this layer.}
            \begin{center}
            \begin{tabular}{lcccc}
                \toprule
              &  \(\tau=0.1, \eta=0.1\)& \(\tau=0.01, \eta=0.1\)& \(\tau=0.1, \eta=1\) & \(\tau=0.01, \eta=1\)
            \\  
            \midrule 
            Clean Acc&  90.52 & 91.65 & 88.88 & 91.14 \\ 
            \(\cW_{\infty}\) Adversarial Acc & 68.26 & 70.04 & \textbf{68.71} & 70.26\\
            \(\cW_{2}\) Adversarial Acc& 54.71 & 56.84 & \textbf{58.02} & 57.46\\
            \bottomrule
            \end{tabular}
            \end{center}
    \end{table*}

            \begin{table*}[htbp]
            \caption{Performance of  Cui et al. (2023) under different configurations. We perturbed the whole network with a small noise and trained all the layers.}
            \begin{center}
            \begin{tabular}{lcccc}
                \toprule
              &  \(\tau=0.1, \eta=0.1\)& \(\tau=0.01, \eta=0.1\)& \(\tau=0.1, \eta=1\) & \(\tau=0.01, \eta=1\)
            \\  
            \midrule 
            Clean Acc& 64.43 & 88.10 & 57.68 & 86.27 \\ 
            \(\cW_{\infty}\) Adversarial Acc & 33.68 & 66.07& 31.12 & 66.64\\
            \(\cW_{2}\) Adversarial Acc&  19.18 & 51.10 & 18.69 & 55.48\\
            \bottomrule
            \end{tabular}
            \end{center}
    \end{table*}

          \begin{table*}[htbp]
            \caption{Performance of Wang et al. (2023) under different configurations. We randomized the last layer and fine-tuned the network by only training this layer.}
            \begin{center}
            \begin{tabular}{lcccc}
                \toprule
              &  \(\tau=0.1, \eta=0.1\)& \(\tau=0.01, \eta=0.1\)& \(\tau=0.1, \eta=1\) & \(\tau=0.01, \eta=1\)
            \\  
            \midrule 
            Clean Acc&  90.71 & 91.48 & 91.45 & 90.67 \\ 
            \(\cW_{\infty}\) Adversarial Acc & 68.94 & 68.94 & \textbf{69.19} & 68.56\\
            \(\cW_{2}\) Adversarial Acc& 54.38 & 55.49 & \textbf{55.93} & 55.09\\
            \bottomrule
            \end{tabular}
            \end{center}
    \end{table*}

            \begin{table*}[htbp]
            \caption{Performance of  Wang et al. (2023) under different configurations. We perturbed the whole network with a small noise and trained all the layers.}
            \begin{center}
            \begin{tabular}{lcccc}
                \toprule
              &  \(\tau=0.1, \eta=0.1\)& \(\tau=0.01, \eta=0.1\)& \(\tau=0.1, \eta=1\) & \(\tau=0.01, \eta=1\)
            \\  
            \midrule 
            Clean Acc& 66.10 & 87.97 & 62.94 & 84.45 \\ 
            \(\cW_{\infty}\) Adversarial Acc & 35.74 & 62.57& 35.43 & 60.44\\
            \(\cW_{2}\) Adversarial Acc&  23.10 & 47.32 & 23.26 & 49.00\\
            \bottomrule
            \end{tabular}
            \end{center}
    \end{table*}

\end{document}